# An improved computer vision method for detecting white blood cells

**Erik Cuevas[1], Margarita Díaz, Miguel Manzanares, Daniel Zaldivar and Marco Pérez**

[*]Departamento de Electrónica
Universidad de Guadalajara, CUCEI
Av. Revolución 1500, Guadalajara, Jal, México
[1]erik.cuevas@cucei.udg.mx

## Abstract

The automatic detection of White Blood Cells (WBC) still remains as an unsolved issue in medical imaging. The analysis of WBC images has engaged researchers from fields of medicine and computer vision alike. Since WBC can be approximated by an ellipsoid form, an ellipse detector algorithm may be successfully applied in order to recognize them. This paper presents an algorithm for the automatic detection of WBC embedded into complicated and cluttered smear images that considers the complete process as a multi-ellipse detection problem. The approach, based on the Differential Evolution (DE) algorithm, transforms the detection task into an optimization problem where individuals emulate candidate ellipses. An objective function evaluates if such candidate ellipses are really present in the edge image of the smear. Guided by the values of such function, the set of encoded candidate ellipses (individuals) are evolved using the DE algorithm so that they can fit into the WBC enclosed within the edge-only map of the image. Experimental results from white blood cell images with a varying range of complexity are included to validate the efficiency of the proposed technique in terms of accuracy and robustness.

## 1. Introduction

Medical image processing has become more and more important in diagnosis with the development of medical imaging and computer technique. Huge amounts of medical images are obtained by X-ray radiography, CT and MRI. They provide essential information for efficient and accurate diagnosis based on advance computer vision techniques [1,2].

On the other hand, White Blood Cells (WBC) also known as leukocytes play a significant role in the diagnosis of different diseases. Although computer vision techniques have successfully contributed to generate new methods for cell analysis, which in turn, have lead into more accurate and reliable systems for disease diagnosis. However, high variability on cell shape, size, edge and localization, complicates the data extraction process. Moreover, the contrast between cell boundaries and the image's background may vary due to unstable lighting conditions during the capturing process.

Many works have been conducted in the area of blood cell detection. In [3] a method based on boundary support vectors is proposed to identify WBC. In such approach, the intensity of each pixel is used to construct feature vectors whereas a Support Vector Machine (SVM) is used for classification and segmentation. By using a different approach, in [4], Wu et. al developed an iterative Otsu method based on the circular histogram for leukocyte segmentation. According to such technique, the smear images are processed in the Hue-Saturation-Intensity (HSI) space by considering that the Hue component contains most of the WBC information. One of the latest advances in white blood cell detection research is the algorithm proposed by Wang [5] that is based on the fuzzy cellular neural network (FCNN). Although such method has proved successful in detecting only one leukocyte in the image, it has not been tested over images containing several white cells. Moreover, its performance commonly decays when the iteration number is not properly defined, yielding a challenging problem itself with no clear clues on how to make the best choice.

---

[1] Corresponding author, Tel +52 33 1378 5900,  ext. 27714, E-mail: erik.cuevas@cucei.udg.mx

Since white blood cells can be approximated with an ellipsoid form, computer vision techniques for detecting ellipses may be used in order to recognize them. Ellipse detection in real images is an open research problem since long time ago. Several approaches have been proposed which traditionally fall under three categories: Symmetry-based, Hough transform-based (HT) and Random sampling.

In symmetry-based detection [6,7], the ellipse geometry is taken into account. The most common elements used in ellipse geometry are the ellipse center and axis. Using these elements and edges in the image, the ellipse parameters can be found. Ellipse detection in digital images is commonly solved through the Hough Transform [8]. It works by representing the geometric shape by its set of parameters, then accumulating bins in the quantized parameter space. Peaks in the bins provide the indication of where ellipses may be. Obviously, since the parameters are quantized into discrete bins, the intervals of the bins directly affect the accuracy of the results and the computational effort. Therefore, for fine quantization of the space, the algorithm returns more accurate results, while suffering from large memory loads and expensive computation. In order to overcome such a problem, some other researchers have proposed other ellipse detectors following the Hough transform principles by using random sampling. In random sampling-based approaches [9,10], a bin represents a candidate shape rather than a set of quantized parameters, as in the HT. However, like the HT, random sampling approaches go through an accumulation process for the bins. The bin with the highest score represents the best approximation of an actual ellipse in the target image. McLaughlin's work [11] shows that a random sampling-based approach produces improvements in accuracy and computational complexity, as well as a reduction in the number of false positives (non existent ellipses), when compared to the original HT and the number of its improved variants.

As an alternative to traditional techniques, the problem of ellipse detection has also been handled through optimization methods. In general, they have demonstrated to give better results than those based on the HT and random sampling with respect to accuracy and robustness [12]. Such approaches have produced several robust ellipse detectors using different optimization algorithms such as Genetic algorithms (GA) [13,14] and Particle Swarm Optimization (PSO) [15].

Although detection algorithms based on optimization approaches present several advantages in comparison to traditional approaches, they have been scarcely applied to WBC detection. One exception is the work presented by Karkavitsas & Rangoussi [16] that solves the WBC detection problem through the use of GA. However, since the evaluation function, which assesses the quality of each solution, considers the number of pixels contained inside of a circle with fixed radius, the method is prone to produce misdetections particularly for images that contained overlapped or irregular WBC.

In this paper, the WBC detection task is approached as an optimization problem and the differential evolution algorithm is used to build the ellipsoidal approximation. Differential Evolution (DE), introduced by Storn and Price [27], is a novel evolutionary algorithm which is used to optimize complex continuous nonlinear functions. As a population-based algorithm, DE uses simple mutation and crossover operators to generate new candidate solutions, and applies one-to-one competition scheme to greedily decide whether the new candidate or its parent will survive in the next generation. Due to its simplicity, ease of implementation, fast convergence, and robustness, the DE algorithm has gained much attention, reporting a wide range of successful applications in the literature [18-22].

This paper presents an algorithm for the automatic detection of blood cell images based on the DE algorithm. The proposed method uses the encoding of five edge points as candidate ellipses in the edge map of the smear. An objective function allows to accurately measure the resemblance of a candidate ellipse with an actual WBC on the image. Guided by the values of such objective function, the set of encoded candidate ellipses are evolved using the DE algorithm so that they can fit into actual WBC on the image. The approach generates a sub-pixel detector which can effectively identify leukocytes in real images. Experimental evidence shows the effectiveness of such method in detecting leukocytes despite complex conditions. Comparison to the state-of-the-art WBC detectors on multiple images demonstrates a better performance of the proposed method.

The main contribution of this study is the proposal of a new WBC detector algorithm that efficiently recognize WBC under different complex conditions while considering the whole process as an ellipse detection problem. Although ellipse detectors based on optimization present several interesting properties, to the best of our knowledge, they have not yet been applied to any medical image processing up to date.

This paper is organized as follows: Section 2 provides a description of the DE algorithm while in Section 3 the ellipse detection task is fully explained from an optimization perspective within the context of the DE approach. The complete WBC detector is presented in Section 4. Section 5 reports the obtained experimental results whereas Section 6 conducts a comparison between state-of-the-art WBC detectors and the proposed approach. Finally, in section 7, some conclusions are drawn.

## 2. Differential evolution algorithm

The DE algorithm is a simple and direct search algorithm which is based on population and aims for optimizing global multi-modal functions. DE employs the mutation operator as to provide the exchange of information among several solutions.

There are various mutation base generators to define the algorithm type. The version of DE algorithm used in this work is known as rand-to-best/1/bin or ''DE1'' [17]. DE algorithms begin by initializing a population of $N_p$ and $D$-dimensional vectors considering parameter values that are randomly distributed between the pre-specified lower initial parameter bound $x_{j,\text{low}}$ and the upper initial parameter bound $x_{j,\text{high}}$ as follows:

$$x_{j,i,t} = x_{j,\text{low}} + \text{rand}(0,1) \cdot (x_{j,\text{high}} - x_{j,\text{low}});$$
$$j = 1, 2, \ldots, D; \quad i = 1, 2, \ldots, N_p; \quad t = 0. \tag{1}$$

The subscript $t$ is the generation index, while $j$ and $i$ are the parameter and particle indexes respectively. Hence, $x_{j,i,t}$ is the $j$th parameter of the $i$th particle in generation $t$. In order to generate a trial solution, DE algorithm first mutates the best solution vector $\mathbf{x}_{best,t}$ from the current population by adding the scaled difference of two vectors from the current population.

$$\mathbf{v}_{i,t} = \mathbf{x}_{best,t} + F \cdot (\mathbf{x}_{r_1,t} - \mathbf{x}_{r_2,t});$$
$$r_1, r_2 \in \left\{ 1, 2, \ldots, N_p \right\} \tag{2}$$

with $\mathbf{v}_{i,t}$ being the mutant vector. Indices $r_1$ and $r_2$ are randomly selected with the condition that they are different and have no relation to the particle index $i$ whatsoever (i.e., $r_1 \neq r_2 \neq i$). The mutation scale factor $F$ is a positive real number, typically less than one. Figure 1 illustrates the vector-generation process defined by Equation (2).

In order to increase the diversity of the parameter vector, the crossover operation is applied between the mutant vector $\mathbf{v}_{i,t}$ and the original individuals $\mathbf{x}_{i,t}$. The result is the trial vector $\mathbf{u}_{i,t}$ which is computed by considering element to element as follows:

$$u_{j,i,t} = \begin{cases} v_{j,i,t}, & \text{if rand}(0,1) \leq CR \text{ or } j = j_{\text{rand}}, \\ x_{j,i,t}, & \text{otherwise.} \end{cases} \tag{3}$$

with $j_{\text{rand}} \in \left\{ 1, 2, \ldots, D \right\}$. The crossover parameter $(0.0 \leq CR \leq 1.0)$ controls the fraction of parameters that the mutant vector is contributing to the final trial vector. In addition, the trial vector always inherits the mutant vector parameter according to the randomly chosen index $j_{\text{rand}}$, assuring that the trial vector differs by at least one parameter from the vector to which it is compared ($\mathbf{x}_{i,t}$).

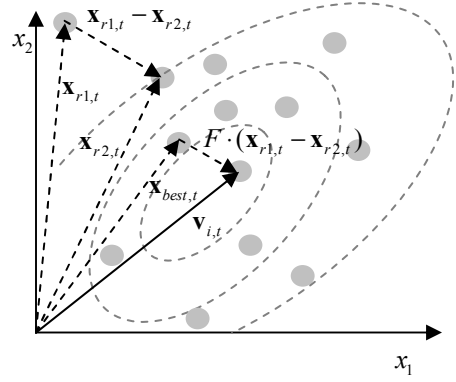

**Fig. 1.** Two-dimensional example of an objective function showing its contour lines and the process for generating $\mathbf{v}$ in scheme DE/best/l/exp from vectors of the current generation.

Finally, a greedy selection is used to find better solutions. Thus, if the computed cost function value of the trial vector $\mathbf{u}_{i,t}$ is less or equal than the cost of the vector $\mathbf{x}_{i,t}$, then such trial vector replaces $\mathbf{x}_{i,t}$ in the next generation. Otherwise, $\mathbf{x}_{i,t}$ remains in the population for at least one more generation:

$$\mathbf{x}_{i,t+1} = \begin{cases} \mathbf{u}_{i,t}, & \text{if } f(\mathbf{u}_{i,t}) \leq f(\mathbf{x}_{i,t}), \\ \mathbf{x}_{i,t}, & \text{otherwise.} \end{cases} \tag{4}$$

Here, $f()$ represents the objective function. These processes are repeated until a termination criterion is attained or a predetermined generation number is reached.

### 3. Ellipse detection using DE.

#### 3.1. Data preprocessing

In order to detect ellipse shapes, candidate images must be preprocessed first by an edge detection algorithm which yields an edge map image. Then, the $(x_i, y_i)$ coordinates for each edge pixel $p_i$ are stored inside the edge vector $P = \left\{ p_1, p_2, \ldots, p_{N_p} \right\}$, with $N_p$ being the total number of edge pixels.

#### 3.2. Individual representation

Similar to lines which need two different points to define them; ellipses require five points to draw one. Thus, each candidate solution $E$ (ellipse candidate) uses five edge points to encode an individual. Under such representation, edge points are selected following a random positional index within the edge array $P$. This procedure will encode a candidate solution as the ellipse that passes through five points $p_1$, $p_2$, $p_3$, $p_4$ and $p_5$ ($E = \{p_1, p_2, p_3, p_4, p_5\}$). Thus, by substituting the coordinates of each point of $E$ into Eq. 5, we gather a set of five simultaneous equations which are linear in the five unknown parameters $a, b, f, g$ and $h$.

$$ax^2 + 2hxy + by^2 + 2gx + 2fy + 1 = 0 \tag{5}$$

Considering the configuration of the edge points shown by Figure 2, the ellipse center $(x_0, y_0)$, the radius maximum ($r_{max}$), the radius minimum ($r_{min}$) and the ellipse orientation ($\theta$) can be calculated as follows:

$$x_0 = \frac{hf - bg}{C}, \qquad (6)$$

$$y_0 = \frac{gh - af}{C}, \qquad (7)$$

$$r_{max} = \sqrt{\frac{-2\Delta}{C(a + b - R)}}, \qquad (8)$$

$$r_{min} = \sqrt{\frac{-2\Delta}{C(a + b + R)}}, \qquad (9)$$

$$\theta = \frac{1}{2}\arctan\left(\frac{2h}{a - b}\right) \qquad (10)$$

where

$$R^2 = (a - b)^2 + 4h^2, \quad C = ab - h^2 \ \text{ and } \ \Delta = \det\begin{pmatrix} \begin{vmatrix} a & h & g \\ h & b & f \\ g & f & 1 \end{vmatrix} \end{pmatrix} \qquad (11)$$

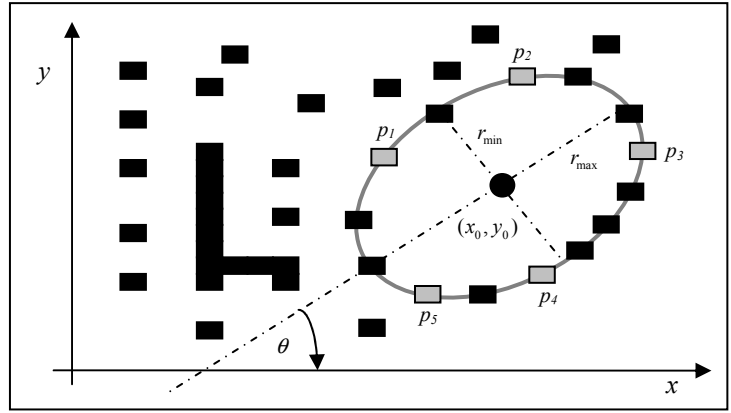

**Fig. 2.** Ellipse candidate (individual) built from the combination of points $p_1$, $p_2$, $p_3$, $p_4$ and $p_5$.

### 3.3 Objective function

Optimization refers to choosing the best element from one set of available alternatives. In the simplest case, it means to minimize an objective function or error by systematically choosing the values of variables from their valid ranges. In order to calculate the error produced by a candidate solution $E$, the ellipse coordinates are calculated as a virtual shape which, in turn, must also be validated, i.e. if it really exists in the edge image. The test set is represented by $S = \{s_1, s_2, \ldots, s_{N_s}\}$, where $N_s$ are the number of points over which the existence of an edge point, corresponding to $E$, should be tested.

The set $S$ is generated by the Midpoint Ellipse Algorithm (MEA) [23] which is a searching method that seeks required points for drawing an ellipse. Any point $(x, y)$ on the boundary of the ellipse with $a,h,b,g$ and $f$ satisfies the equation $f_{ellipse}(x,y) \cong r_{max}x^2 + r_{min}y^2 - r_{max}{}^2 r_{min}{}^2$, where $r_{max}$ and $r_{min}$ represent the major and minor axis, respectively. However, MEA avoids computing square-root calculations by comparing the pixel separation distances. A method for direct distance comparison is to test the halfway position between two pixels (sub-pixel distance) to determine if this midpoint is inside or outside the ellipse boundary. If the point is in the interior of the ellipse, the ellipse function is negative. Thus, if the point is outside the ellipse, the ellipse function is positive. Therefore, the error involved in locating pixel positions using the midpoint test is limited to one-half the pixel separation (sub-pixel precision). To summarize, the relative position of any point $(x, y)$ can be determined by checking the sign of the ellipse function:

$$f_{ellipse}(x,y) \begin{cases} < 0 & \text{if } (x, y) \text{ is inside the ellipse boundary} \\ = 0 & \text{if } (x, y) \text{ is on the ellipse boundary} \\ > 0 & \text{if } (x, y) \text{ is outside the ellipse boundary} \end{cases} \tag{12}$$

The ellipse-function test in Eq. 12 is applied to mid-positions between pixels nearby the ellipse path at each sampling step. Figure 3a and 4a show the midpoint between the two candidate pixels at sampling position. The ellipse is used to divide the quadrants into two regions the limit of the two regions is the point at which the curve has a slope of -1 as shown in Figure 4.

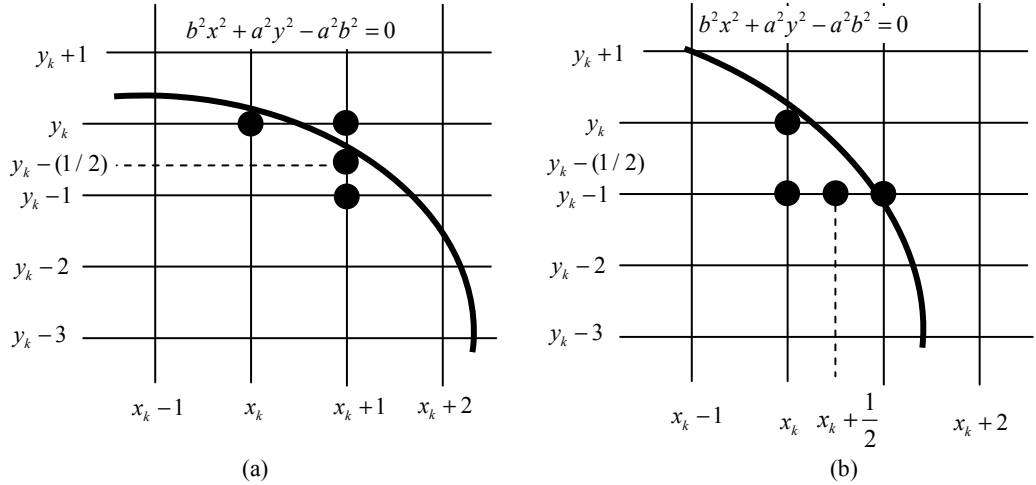

(a)          (b)

**Fig. 3.** (a) symmetry of the ellipse: an estimated one octant which belong to the first region where the slope is greater than -1, b) In this region the slope will be less than -1 to complete the octant and continue to calculate the same so the remaining octants.

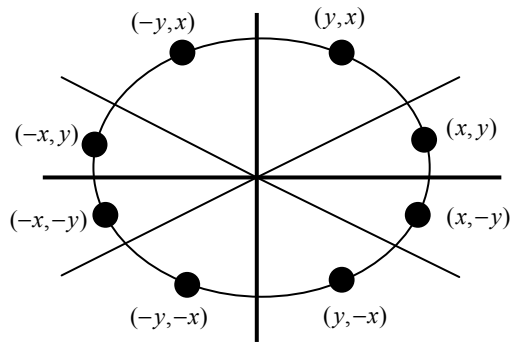

**Fig. 4.** Midpoint between candidate pixels at sampling position $x_k$ along an elliptical path.

In MEA the computation time is reduced by considering the symmetry of ellipses. Ellipses sections in adjacent octants within one quadrant are symmetric with respect to the dy/dy=-1 line dividing the two octants. These symmetry conditions are illustrated in Figure 4. The algorithm can be considered as the quickest providing a sub-pixel precision [24]. However, in order to protect the MEA operation, it is important to assure that points lying outside the image plane must not be considered in S.

The objective function $J(E)$ represents the matching error produced between the pixels $S$ of the ellipse candidate $E$ and the pixels that actually exist in the edge image, yielding:

$$J(E) = 1 - \frac{\sum_{v=1}^{Ns} G(x_v, y_v)}{Ns} \tag{13}$$

where $G(x_i, y_i)$ is a function that verifies the pixel existence in $(x_v, y_v)$, with $(x_v, y_v) \in S$ and $N_x$ being the number of pixels lying on the perimeter corresponding to $E$ currently under testing. Hence, function $G(x_v, y_v)$ is defined as:

$$G(x_v, y_v) = \begin{cases} 1 & \text{if the pixel } (x_v, y_v) \text{ is an edge point} \\ 0 & \text{otherwise} \end{cases} \tag{14}$$

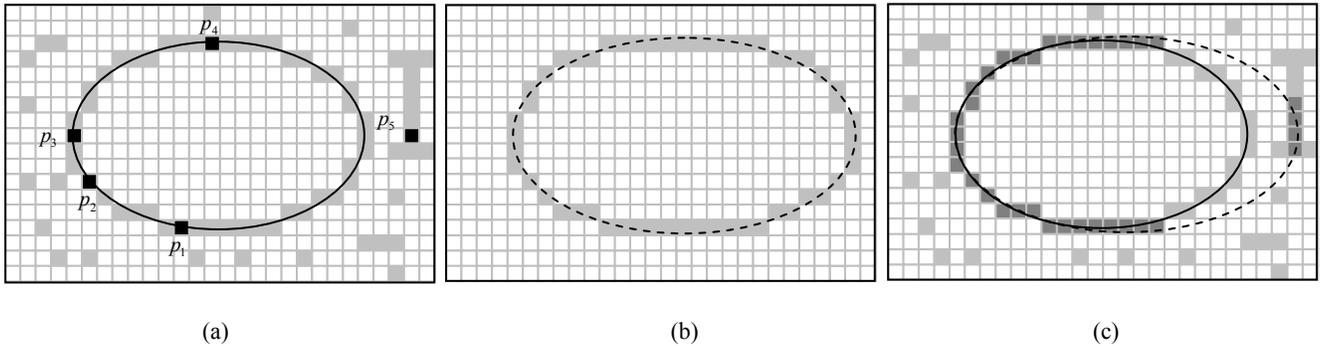

**Fig. 5.** Evaluation of a candidate solution $E$: the image in (a) shows the original image while (b) presents the generated virtual shape drawn from points $p_1$, $p_2$, $p_3$, $p_4$ and $p_5$. The image in (c) shows coincidences between both images which have been marked by darker pixels while the virtual shape is also depicted through a dashed line

A value of $J(E)$ near to zero implies a better response from the "ellipsoid" operator. Figure 5 shows the procedure to evaluate a candidate action $E$ with its representation as a virtual shape $S$. Figure 5(a) shows the original edge map, while Figure 5(b) presents the virtual shape $S$ representing the individual $E = \{p_1, p_2, p_3, p_4, p_5\}$. In Figure 5(c), the virtual shape $S$ is compared to the original image, point by point, in order to find coincidences between virtual and edge points. The individual has been built from points $p_1$, $p_2$, $p_3$, $p_4$ and $p_5$ which are shown by Fig. 5(a). The virtual shape $S$, obtained by MEA, gathers 52 points ($N_s$ = 52) with only 35 of them existing in both images (shown as darker points in Fig. 5(c)) and yielding: $\sum_{v=1}^{Ns} G(x_v, y_v) = 35$, therefore $J(E)$=0.327.

### 3.4 Implementation of DE for ellipse detection

The ellipse detector algorithm based on DE can be summarized in the following steps:

Step 1:    Set the DE parameters $F$=0.25 and $CR$=0.8.

Step 2:    Initialize the population of $m$ individuals $\mathbf{E}^k = \{E_1^k, E_2^k, \ldots, E_m^k\}$ where each decision variable $p_1$, $p_2$, $p_3$, $p_4$ and $p_5$ of $E_a^k$ is set randomly within the interval $[1, N_p]$. All values must be integers. Considering that $k$=0 and $a \in (1, 2, \ldots, m)$ .

Step 3:    Evaluate the objective value $J(E_a^k)$ for all $m$ individuals, and determining the $E^{best,k}$ showing the best fitness value, such that $E^{best,k} \in \{\mathbf{E}^k\} \Big| J(E^{best,k}) = \min\{J(E_1^k), J(E_2^k), \ldots, J(E_m^k)\}$ .

Step 4:    Generate the trial population $\mathbf{T} = \{T_1, T_2, \ldots, T_m\}$ :

for ($i$=1; $i$<$m$+1; $i$++)
do $r_1$=floor(rand(0,1)·$m$ ); while ( $r_1 = i$ );
do $r_2$=floor(rand(0,1)·$m$ ); while (( $r_2 = i$ ) or ( $r_2 = r_2$ ));
jrand=floor(5·rand(0,1));

    for ($j$=1; $j$<6; $j$++) // generate a trial vector
      if (rand(0,1)<=$CR$ or $j$=jrand)
      $T_{j,i} = E_j^{best,k} + F \cdot (E_{j,r_1}^k - E_{j,r_2}^k)$;
        else
          $T_{j,i} = E_{j,i}^k$;
      end if

    end for

end for

Step 5:    Evaluate the fitness values $J(T_i)$ ($i \in \{1, 2, \ldots, m\}$) of all trial individuals. Check all individuals. If a candidate parameter set is not physically plausible, i.e. out of the range $[1, N_p]$, then an exaggerated cost function value is returned. This aims to eliminate ''unstable'' individuals.

Step 6:    Select the next population $\mathbf{E}^{k+1} = \{E_1^{k+1}, E_2^{k+1}, \ldots, E_m^{k+1}\}$ :

for ($i$=1; $i$<$m$+1; $i$++)
    if $(J(T_i) < J(E_i^k))$
    $E_i^{k+1} = T_i$
      else
        $E_i^{k+1} = E_i^k$
    end if
end for

Step 7:    If the iteration number ($NI$) is met, then the output $E^{best,k}$ is the solution (an actual ellipse contained in the image), otherwise go back to Step 3.

## 4. The White blood cell detector

In order to detect WBC, the proposed detector combines a segmentation strategy with the ellipse detection approach presented in section 3.

### 4.1 Image preprocessing

To employ the proposed detector, smear images must be preprocessed to obtain two new images: the segmented image and its corresponding edge map. The segmented image is produced by using a segmentation strategy whereas the edge map is generated by a border extractor algorithm. Such edge map is considered by the objective function to measure the resemblance of a candidate ellipse with an actual WBC.

The goal of the segmentation strategy is to isolate the white blood cells (WBC's) from other structures such as red blood cells and background pixels. Information of color, brightness and gradients are commonly used within a thresholding scheme to generate the labels to classify each pixel. Although a simple histogram thresholding can be used to segment the WBC's, at this work the Diffused Expectation-Maximization (DEM) has been used to assure better results [25].

DEM is an Expectation-Maximization (EM) based algorithm which has been used to segment complex medical images [26]. In contrast to classical EM algorithms, DEM considers the spatial correlations among pixels as a part of the minimization criteria. Such adaptation allows to segment objects in spite of noisy and complex conditions. The method models an image as a finite mixture, where each mixture component corresponds to a region class and uses a maximum likelihood approach to estimate the parameters of each class, via the expectation maximization (EM) algorithm, coupled with anisotropic diffusion on classes, in order to account for the spatial dependencies among pixels.

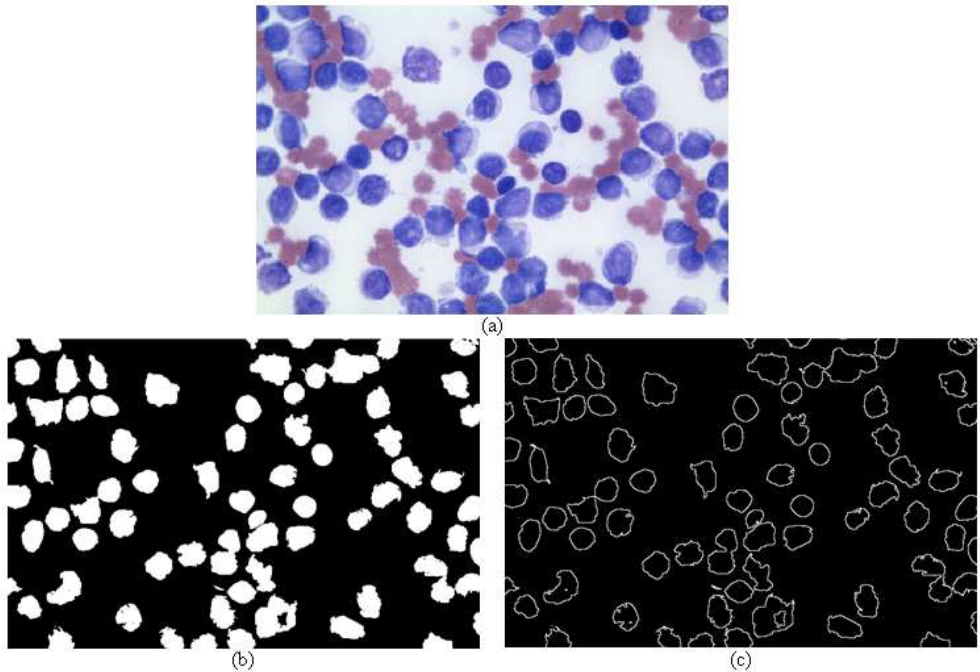

**Fig. 6.** Preprocessing process. (a) original smear image, (b) segmented image obtained by DEM and (c) the edge map obtained by using the morphological edge detection procedure.

For the WBC's segmentation, it has been used the implementation of DEM provided in [27]. Since the implementation allows to segment gray-level images and color images, it can be used for operating over all smear images without matter the way in which they are acquired. The DEM has been configured considering three different classes ($K$=3), $g\left(\nabla h_{ik}\right)=\left|\nabla h_{ik}\right|^{-9/5}$, $\lambda = 0.1$ and $m$=10 iterations. These values have been found as the best configuration set according to [25].

As a final result of the DEM operation, three different thresholding points are obtained: the first corresponds to the WBC's, the second to the red blood cells whereas the third represents the pixels classified as background. Figure 6(b) presents the segmentation results obtained by the DEM approach employed at this work considering the Figure 6(a) as the original image.

Once the segmented image has been produced, the edge map is computed. The purpose of the edge map is to obtain a simple image representation that preserves object structures. The DE-based detector operates directly over the edge map in order to recognize ellipsoidal shapes. Several algorithms can be used to extract the edge map; however, at this work, the morphological edge detection procedure [28] has been used to accomplish such a task. Morphological edge detection is a traditional method to extract borders from binary images in which original images ($I_B$) are eroded by a simple structure element ($I_E$) composed by a template of 3x3 ones. Then, the eroded image is inverted ($\overline{I}_E$) and compared with the original image ($\overline{I}_E \wedge I_B$) in order to detect pixels which are present in both images. Such pixels compose the computed edge map from $I_B$. Figure 6(c) shows the edge map obtained by using the morphological edge detection procedure.

### 4.2 Ellipse detection approach

The edge map is used as input image for the ellipse detector presented in Section 3. After several calibration experiments, Table 1 presents the parameters used in this work for the DE algorithm. The final configuration coincides with the best possible calibration proposed in [29], where it has been analyzed the effect of modifying the DE-parameters in several generic optimization problems. The population-size parameter ($m$=20) has been selected considering the best possible balance between convergence and computational overload. Once defined, such configuration has been kept for all test images employed in the experimental study.

| $m$ | $F$ | $CR$ | $NI$ |
|-----|------|------|------|
| 20 | 0.25 | 0.80 | 200 |

**Table 1.** DE parameters used for leukocites detection in medical images.

Under such assumptions, the complete process to detect WBC's is implemented as follows:

**Step 1:**      Segment the WBC's using the DEM algorithm (described in 4.1)
**Step 2:**      Get the edge map from the segmented image.
**Step 3:**      Start the ellipse detector based in DE over the edge map while saving best ellipses (Section 3).
**Step 4:**      Define parameter values for each ellipse that identify the WBC's.

### 4.3 Numerical example

In order to present the algorithm's step-by-step operation, a numerical example has been set by applying the proposed method to detect a single leukocyte lying inside of a simple image. Fig. 7(a) shows the image used in the example. After applying the threshold operation, the WBC is located besides few other pixels which are merely noise (see Fig. 7(b)). Then, the edge map is subsequently computed and stored pixel by pixel inside the vector $P$. Fig. 7(c) shows the resulting image after such procedure.

The DE-based ellipse detector is executed using information of the edge map (for the sake of easiness, it only considers a population of four particles). Like all evolutionary approaches, DE is a population-based optimizer that attacks the starting point problem by sampling the search space at multiple, randomly chosen, initial particles. By taking five random pixels from vector $P$, four different particles are constructed. Fig. 7(d) depicts the initial particle distribution $\mathbf{E}^0 = \{E_1^0, E_2^0, E_3^0, E_4^0\}$ . By using the DE operators, four different trial particles $\mathbf{T} = \{T_1, T_2, T_3, T_4\}$ (ellipses) are generated, their locations are shown in Fig. 7(e). Then, the new population $\mathbf{E}^1$ is selected considering the best elements obtained among the trial elements $\mathbf{T}$ and the initial particles $\mathbf{E}^0$ . The final distribution of the new population is depicted in Fig. 7(f). Since the particles $E_2^0$ and $E_2^0$ hold (in Fig. 7(f)) a better fitness value ( $J(E_2^0)$ and $J(E_3^0)$ ) than the trial elements $T_2$ and $T_3$ , they are considered as particles of the final population $\mathbf{E}^1$ . Figures 7(g) and 7(h) present the second iteration produced by the algorithm whereas Fig. 6(i) shows the population configuration after 25 iterations. From Fig. 7(i), it is clear that all particles have converged to a final position which is able to accurately cover the WBC.

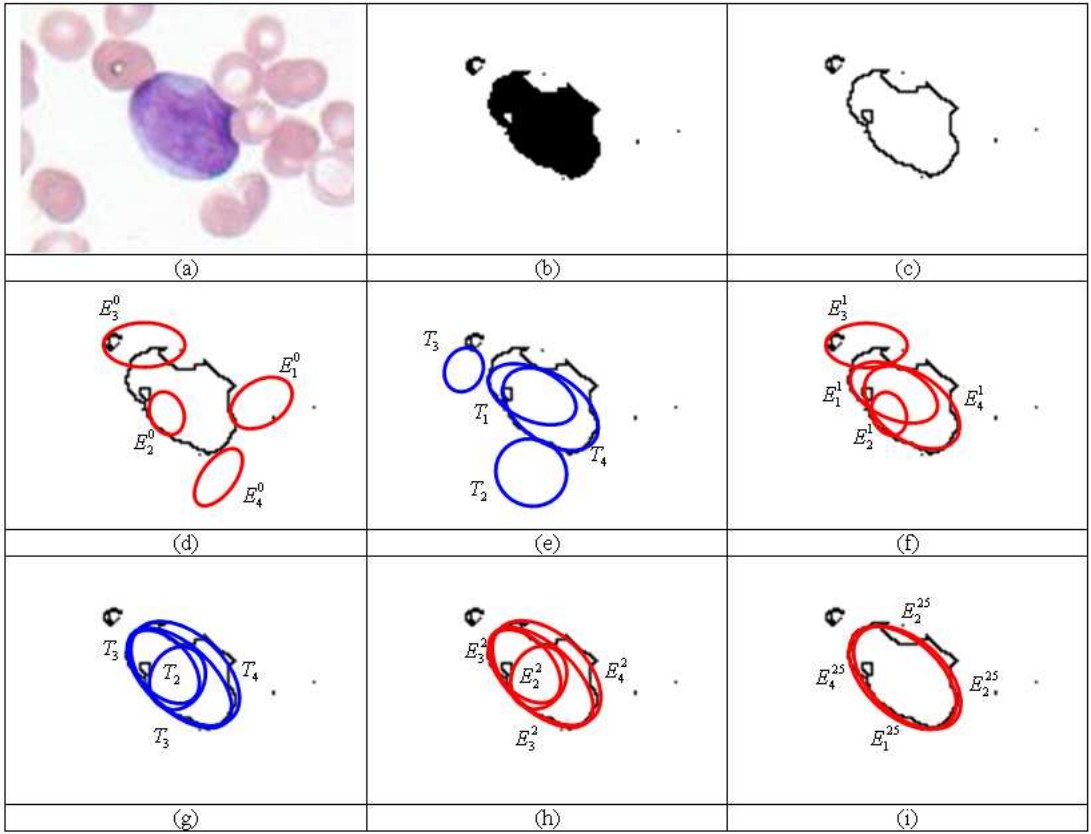

**Fig. 7.** Detection numerical example: (a) The image used as example. (b) Segmented image. (c) Edge map. (d) Initial particles $\mathbf{E}^0$ . (e) Trial elements $\mathbf{T}$ produced by the DE operators. (f) New population $\mathbf{E}^1$ . (g) Trial elements produced considering $\mathbf{E}^1$ as input population. (h) New population $\mathbf{E}^2$ . (i) Final particle configuration after 25 iterations.

## 5. Experimental results

Experimental tests have been developed in order to evaluate the performance of the WBC detector. It was tested over microscope images from blood-smears holding a 960 x 720 pixel resolution. They correspond to supporting images on the leukemia diagnosis. The images show several complex conditions such as deformed cells and overlapping with partial occlusions. The robustness of the algorithm has been tested under such demanding conditions. All the experiments has been developed using a PC based on Intel Core i7-2600, with 8GB in Ram.

Figure 87(a) shows an example image employed in the test. It was used as input image for the WBC detector. Figure 8(b) presents the segmented WBC's obtained by the DEM algorithm. Figures 8(c) and 8(d) present the edge map and the white blood cells after detection, respectively. The results show that the proposed algorithm can effectively detect and mark blood cells despite cell occlusion, deformation or overlapping. Other parameters may also be calculated through the algorithm: the total area covered by white blood cells and relationships between several cell sizes.

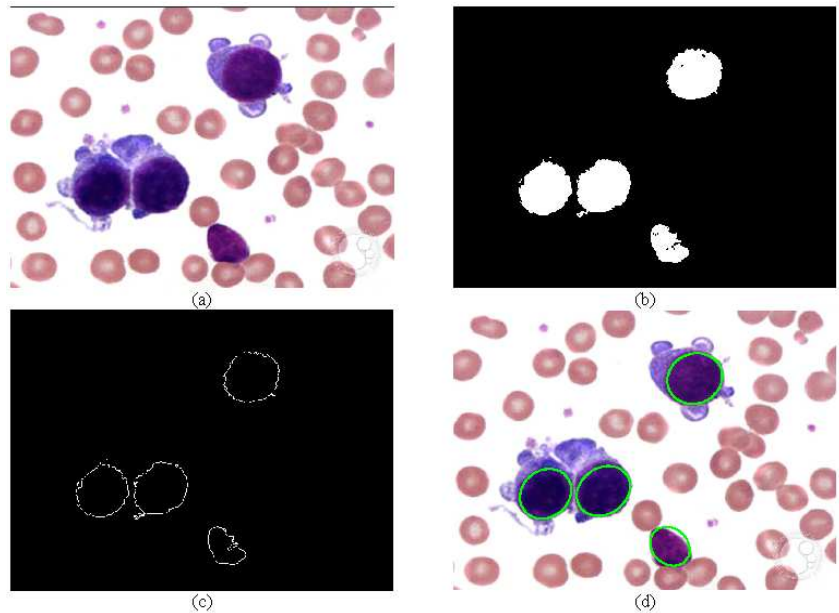

**Fig. 8.** Resulting images of the first test after applying the WBC detector: (a) Original image, (b) image segmented by the DEM algorithm, (c) edge map and (d) the white detected blood cells.

Other example is presented in Figure 9. It represents a complex example with an image showing seriously deformed cells. Despite such imperfections, the proposed approach can effectively detect the cells as it is shown in Figure 9(d).

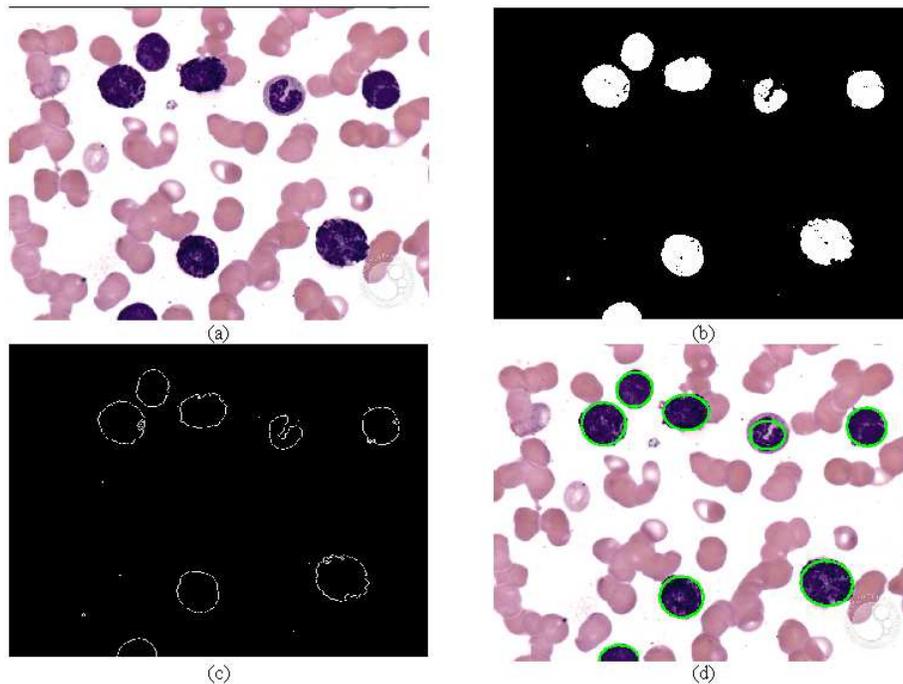

**Fig. 9.** Resulting images of the second test after applying the WBC detector: (a) Original image, (b) image segmented by the DEM algorithm, (c) edge map and (d) the white detected blood cells.

## 6. Comparisons to other methods.

A comprehensive set of smear-blood test images is used to test the performance of the proposed approach. We have applied the proposed DE-based detector to test images in order to compare its performance to other WBC detection algorithms such as the Boundary Support Vectors (BSV) approach [3], the iterative Otsu (IO) method [4], the Wang algorithm [5] and the Genetic algorithm-based (GAB) detector [16]. In all cases, the algorithms are tuned according to the value set which is originally proposed by their own references.

**Fig. 10.** Examples of images included in the experimental set for robustness comparison. (a)-(b) Originals images. (c) Image contaminated with 10% of Salt & Pepper noise and (d) image polluted with $\sigma = 10$ of Gaussian noise.

*6.1 Detection comparison*

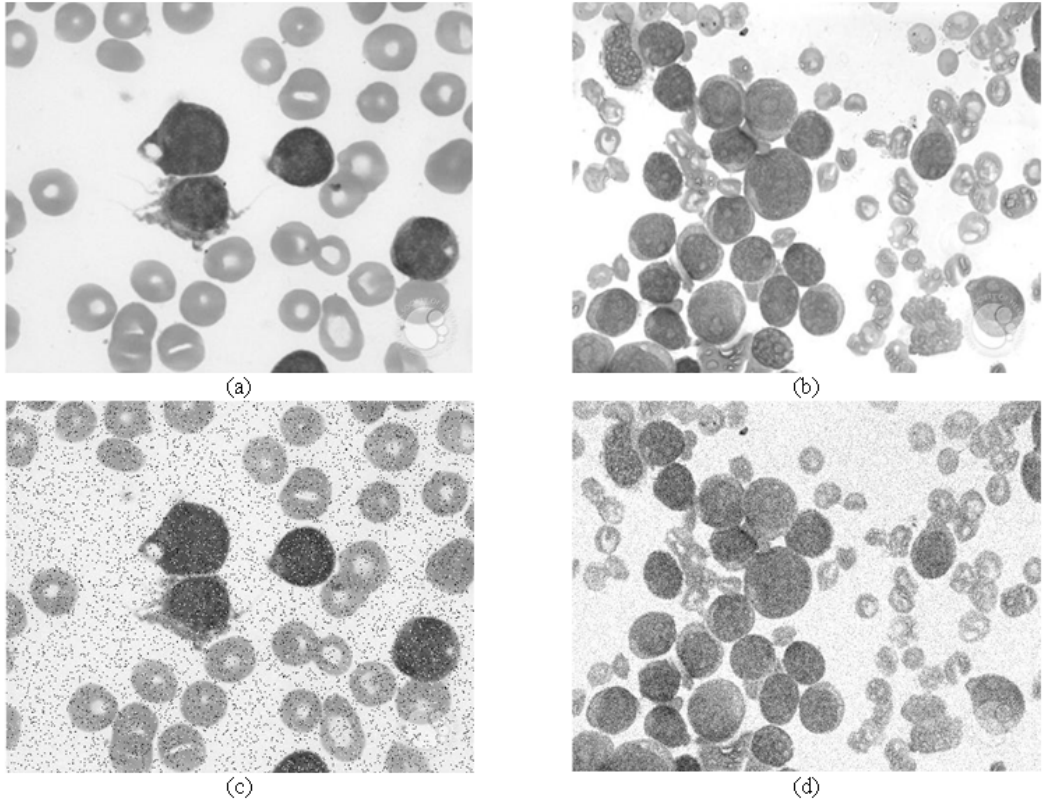

To evaluate the detection performance of the proposed detection method, Table 2 tabulates the comparative leukocyte detection performance of the BSV approach, the IO method, the Wang algorithm, the BGA detector and the proposed method, in terms of detection rates and false alarms. The experimental data set includes 50 images which are collected from the ASH Image Bank (http://imagebank.hematology.org/). Such images contain 517 leukocytes (287 bright leukocytes and 230 dark leukocytes according to smear conditions) which have been detected and counted by a human expert. Such values act as ground truth for all the experiments. For the comparison, the detection rate (DR) is defined as the ratio between the number of leukocytes correctly detected and the number leukocytes determined by the expert. The false alarm rate (FAR) is defined as the ratio between the number of non-leukocyte objects that have been wrongly identified as leukocytes and the number leukocytes which have been actually determined by the expert.

Experimental results show that the proposed DE method, which achieves 98.26% leukocyte detection accuracy with 2.71% false alarm rate, is compared favorably against other WBC detection algorithms, such as the BSV approach, the IO method, the Wang algorithm and the BGA detector.

| Leukocyte Type | Method | Leukocytes detected | Missing | False alarms | DR | FAR |
|---|---|---|---|---|---|---|
| Bright Leukocytes **(287)** | BSV [3] | 130 | 157 | 84 | 45.30% | 29.27% |
| | IO [4] | 227 | 60 | 73 | 79.09% | 25.43% |
| | Wang [5] | 231 | 56 | 60 | 80.49% | 20.90% |
| | GAB [16] | 220 | 67 | 22 | 76.65% | 7.66% |
| | DE-based | 281 | 6 | 11 | 97.91% | 3.83% |
| Dark Leukocytes **(230)** | BSV [3] | 105 | 125 | 59 | 46.65% | 25.65% |
| | IO [4] | 183 | 47 | 61 | 79.56% | 26.52% |
| | Wang [5] | 196 | 34 | 47 | 85.22% | 20.43% |
| | GAB [16] | 179 | 51 | 23 | 77.83% | 10.00% |
| | DE-based | 227 | 3 | 3 | 98.70% | 1.30% |
| Overall **(517)** | BSV [3] | 235 | 282 | 143 | 45.45% | 27.66% |
| | IO [4] | 410 | 107 | 134 | 79.30% | 25.92% |
| | Wang [5] | 427 | 90 | 107 | 82.59% | 20.70% |
| | GAB [16] | 399 | 118 | 45 | 77.18% | 8.70% |
| | DE-based | 508 | 9 | 14 | 98.26% | 2.71% |

**Table 2.** Comparative leukocyte detection performance of the BSV approach, the IO method, the Wang algorithm, the BGA detector and the proposed DE method over the data set which contains 30 images and 426 leukocytes

*6.2 Robustness comparison*

Images of blood smear are often deteriorated by noise due to various sources of interference and other phenomena that affect the measurement processes in imaging and data acquisition systems. Therefore, the detection results depend on the algorithm's ability to cope with different kinds of noises. In order to demonstrate the robustness in the WBC detection, the proposed DE approach is compared to the BSV approach, the IO method, the Wang algorithm and the BGA detector under noisy environments. In the test, two different experiments have been studied. The first inquest explores the performance of each algorithm when the detection task is accomplished over images corrupted by Salt & Pepper noise. The second experiment considers images polluted by Gaussian noise. Salt & Pepper and Gaussian noise are selected for the robustness analysis because they represent the most compatible noise types commonly found in images of blood smear [30]. The comparison considers the complete set of 50 images presented in Section 6.1 containing 517 leukocytes which have been detected and counted by a human expert. The added noise is produced by MatLab©, considering two noise levels of 5% and 10% for Salt & Pepper noise whereas $\sigma = 5$ and $\sigma = 10$ are used for the case of Gaussian noise. Such noise levels, according to [31], correspond to the best trade of between detection difficulty and real existence in medical imaging. Using higher noise levels, the detection process would be unnecessarily complicated without representing a feasible image condition.

| Noise level | Method | Leukocytes detected | Missing | False alarms | DR | FAR |
|---|---|---|---|---|---|---|
| 5% Salt & Pepper noise 517 Leukocytes | BSV [3] | 185 | 332 | 133 | 34.74% | 26.76% |
| | IO [4] | 311 | 206 | 106 | 63.38% | 24.88% |
| | Wang [5] | 250 | 176 | 121 | 58.68% | 27.70% |
| | GAB [16] | 298 | 219 | 135 | 71.83% | 24.18% |
| | DE-based | 482 | 35 | 32 | 91.55% | 7.04% |
| 10% Salt & Pepper noise 517 Leukocytes | BSV [3] | 105 | 412 | 157 | 20.31% | 30.37% |
| | IO [4] | 276 | 241 | 110 | 53.38% | 21.28% |
| | Wang [5] | 214 | 303 | 168 | 41.39% | 32.49% |
| | GAB [16] | 337 | 180 | 98 | 65.18% | 18.95% |
| | DE-based | 463 | 54 | 31 | 89.55% | 5.99% |

**Table 3.** Comparative WBC detection among methods that considers the complete data set of 30 images corrupted by different levels of Salt & Pepper noise

Fig. 10 shows two examples of the experimental set. The outcomes in terms of the detection rate (DR) and the false alarm rate (FAR) are reported for each noise type in Table 3 and Table 4. The results show that the proposed DE algorithm presents the best detection performance, achieving in the worst case a DR of 89.55% and 91.10%, under contaminated conditions of Salt & Pepper and Gaussian noise, respectively. On the other hand, the DE detector possesses the least degradation performance presenting a FAR value of 5.99% and 6.77%.

| Noise level | Method | Leukocytes detected | Missing | False alarms | DR | FAR |
|---|---|---|---|---|---|---|
| $\sigma = 5$ Gaussian noise 517 Leukocytes | BSV [3] | 214 | 303 | 98 | 41.39% | 18.95% |
| | IO [4] | 366 | 151 | 87 | 70.79% | 16.83% |
| | Wang [5] | 358 | 159 | 84 | 69.25% | 16.25% |
| | GAB [16] | 407 | 110 | 76 | 78.72% | 14.70% |
| | DE-based | 487 | 30 | 21 | 94.20% | 4.06% |
| $\sigma = 10$ Gaussian noise 517 Leukocytes | BSV [3] | 162 | 355 | 129 | 31.33% | 24.95% |
| | IO [4] | 331 | 186 | 112 | 64.02% | 21.66% |
| | Wang [5] | 315 | 202 | 124 | 60.93% | 23.98% |
| | GAB [16] | 363 | 154 | 113 | 70.21% | 21.86% |
| | DE-based | 471 | 46 | 35 | 91.10% | 6.77% |

**Table 4.** Comparative WBC detection among methods that considers the complete data set of 30 images corrupted by different levels of Gaussian noise.

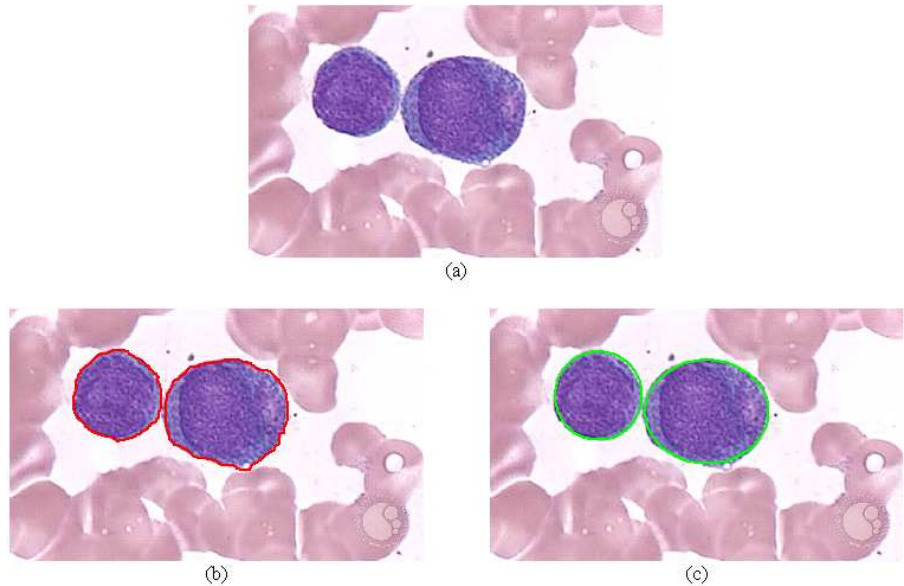

**Fig. 11.** Comparison of the DE and the Wang's method for white blood cell detection in medical images. (a) Original image. (b) Detection using the Wang's method, (c) Detection after applying the DE method.

### 6.3 Stability comparison

In order to compare the stability performance of the proposed method, its results are compared to those reported by Wang *et al.* in [5] which is considered as an accurate technique for the detection of WBC.

The Wang algorithm is an energy-minimizing method which is guided by internal constraint elements and influenced by external image forces, producing the segmentation of WBC's at a closed contour. As external forces, the Wang approach uses edge information which is usually represented by the gradient magnitude of the image. Therefore, the contour is attracted to pixels with large image gradients, i.e. strong edges. At each iteration, the Wang method finds a new contour configuration which minimizes the energy that corresponds to external forces and constraint elements.

In the comparison, the net structure and its operational parameters, corresponding to the Wang algorithm, follow the configuration suggested in [5] while the parameters for the DE-based algorithm are taken from Table 1.

Figure 11 shows the performance of both methods considering a test image with only two white blood cells. Since the Wang method uses gradient information in order to appropriately find a new contour configuration, it needs to be executed iteratively in order to detect each structure (WBC). Figure 11(b) shows the results after the Wang approach has been applied considering only 200 iterations. Furthermore, Figure 11(c) shows results after applying the DE-based method which has been proposed in this paper.

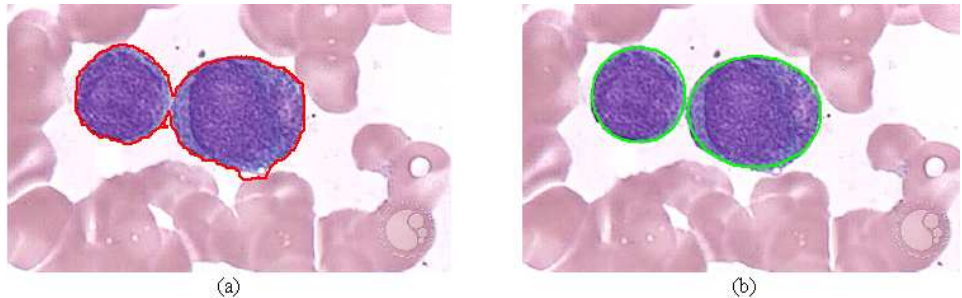

**Fig. 12.** Result comparison for the white blood cells detection showing (a) Wang's algorithm after 400 cycles and (b) DE detector method considering 1000 cycles.

The Wang algorithm uses the fuzzy cellular neural network (FCNN) as optimization approach. It employs gradient information and internal states in order to find a better contour configuration. In each iteration, the FCNN tries, as contour points, different new pixel positions which must be located nearby the original contour position. Such fact might cause the contour solution to remain trapped into a local minimum. In order to avoid such a problem, the Wang method applies a considerable number of iterations so that a near optimal contour configuration can be found. However, when the number of iterations increases the possibility to cover other structures increases too. Thus, if the image has a complex background (just as smear images do) or the WBC's are too close, the method gets confused so that finding the correct contour configuration from the gradient magnitude is not easy. Therefore, a drawback of Wang's method is related to its optimal iteration number (instability). Such number must be determined experimentally as it depends on the image context and its complexity. Figure 12(a) shows the result of applying 400 cycles of the Wang's algorithm while Figure 12(b) presents the detection of the same cell shapes after 1000 iterations using the proposed algorithm. From Fig. 12(a), it can be seen that the contour produced by Wang´s algorithm degenerates as the iteration process continues, wrongly covering other shapes lying nearby.

In order to compare the accuracy of both methods, the estimated WBC area which has been approximated by both approaches, is compared to the actual WBC size considering different degrees of evolution i.e. the cycle number for each algorithm. The comparison considers only one WBC because it is the only detected shape in the Wang's method. Table 5 shows the averaged results over twenty repetitions for each experiment. In order to enhance the result analysis, Fig. 13 presents the response Error % vs. Iterations of an extended version of the outcomes exposed in Table 5

| Algorithm | Iterations | Error% |
|-----------|-----------|--------|
| Wang      | 30        | 88%    |
|           | 60        | 70%    |
|           | 200       | 1%     |
|           | 400       | 121%   |
|           | 600       | 157%   |
| DE-based  | 30        | 24.30% |
|           | 60        | 7.17%  |
|           | 200       | 2.25%  |
|           | 400       | 2.25%  |
|           | 600       | 2.25%  |

**Table 5.** Error in cell's size estimation after applying the DE algorithm and the Wang's method to detect one leukocite embedded into a blood-smear image. The error is averaged over twenty experiments.

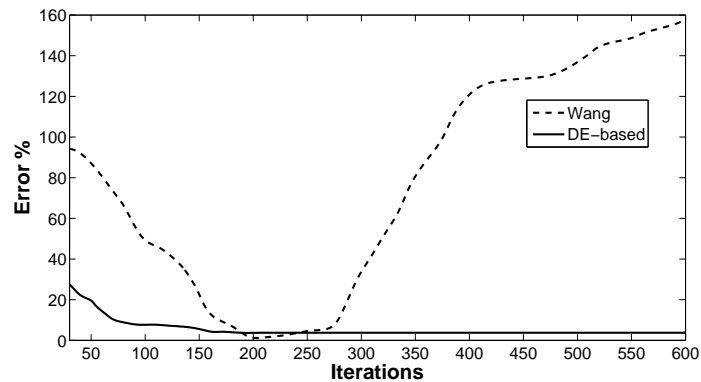

**Fig. 13.** Error % vs. Iterations of an extended version of the results exposed in Table 5

## 7. Conclusions.

In this paper, an algorithm for the automatic detection of blood cell images based on the DE algorithm has been presented. The approach considers the complete process as a multiple ellipse detection problem. The proposed method uses the encoding of five edge points as candidate ellipses in the edge map of the smear. An objective function allows to accurately measure the resemblance of a candidate ellipse with an actual WBC on the image. Guided by the values of such objective function, the set of encoded candidate ellipses are evolved using the DE algorithm so that they can fit into actual WBC on the image. The approach generates a sub-pixel detector which can effectively identify leukocytes in real images.

The performance of the DE-method has been compared to other existing WBC detectors (the Boundary Support Vectors (BSV) approach [3], the iterative Otsu (IO) method [4], the Wang algorithm [5] and the Genetic algorithm-based (GAB) detector [16] considering several images which exhibit different complexity levels. Experimental results demonstrate the high performance of the proposed method in terms of detection accuracy, robustness and stability.